\def\year{2022}\relax
\documentclass[letterpaper]{article} 

\usepackage{aaai22}  
\usepackage{times}  
\usepackage{helvet}  
\usepackage{courier}  
\usepackage[hyphens]{url}  
\usepackage{graphicx} 
\usepackage{natbib}  
\usepackage{caption} 
\DeclareCaptionStyle{ruled}{labelfont=normalfont,labelsep=colon,strut=off} 
\usepackage{enumitem}
\usepackage{amsmath,amssymb,amsthm}
\usepackage{amsfonts}
\usepackage{multirow}
\usepackage[table,xcdraw]{xcolor} 
\usepackage{subcaption}

\def\onedot{\ifx\@let@token.\else.\null\fi\xspace}

\makeatother

\def\Vec#1{{\boldsymbol{#1}}} 
 
\title{A Tale of Color Variants: \\ Representation and Self-Supervised Learning in Fashion E-Commerce}
\author {
    Ujjal Kr Dutta,
    Sandeep Repakula,
    Maulik Parmar,
    Abhinav Ravi
}
\affiliations {
    Data Sciences-Image Sciences, Myntra\\
    \{ujjal.dutta,sandeep.r,abhinav.ravi\}@myntra.com, maulik2087@gmail.com
}

\begin{document}

\maketitle

\begin{abstract}
In this paper, we address a crucial problem in fashion e-commerce (with respect to customer experience, as well as revenue): color variants identification, i.e., identifying fashion products that match exactly in their design (or style), but only to differ in their color. We propose a generic framework, that leverages deep visual Representation Learning at its heart, to address this problem for our fashion e-commerce platform. Our framework could be trained with supervisory signals in the form of triplets, that are obtained manually. However, it is infeasible to obtain manual annotations for the entire huge collection of data usually present in fashion e-commerce platforms, such as ours, while capturing all the difficult corner cases. But, to our rescue, interestingly we observed that this crucial problem in fashion e-commerce could also be solved by simple color jitter based image augmentation, that recently became widely popular in the contrastive Self-Supervised Learning (SSL) literature, that seeks to learn visual representations without using manual labels. This naturally led to a question in our mind: Could we leverage SSL in our use-case, and still obtain comparable performance to our supervised framework? The answer is, Yes! because, color variant fashion objects are nothing but manifestations of a style, in different colors, and a model trained to be invariant to the color (with, or without supervision), should be able to recognize this! This is what the paper further demonstrates, both qualitatively, and quantitatively, while evaluating a couple of state-of-the-art SSL techniques, and also proposing a novel method.
\end{abstract}

\section{Introduction}
\begin{figure}[t]
  \centering
  \includegraphics[width=0.5\columnwidth]{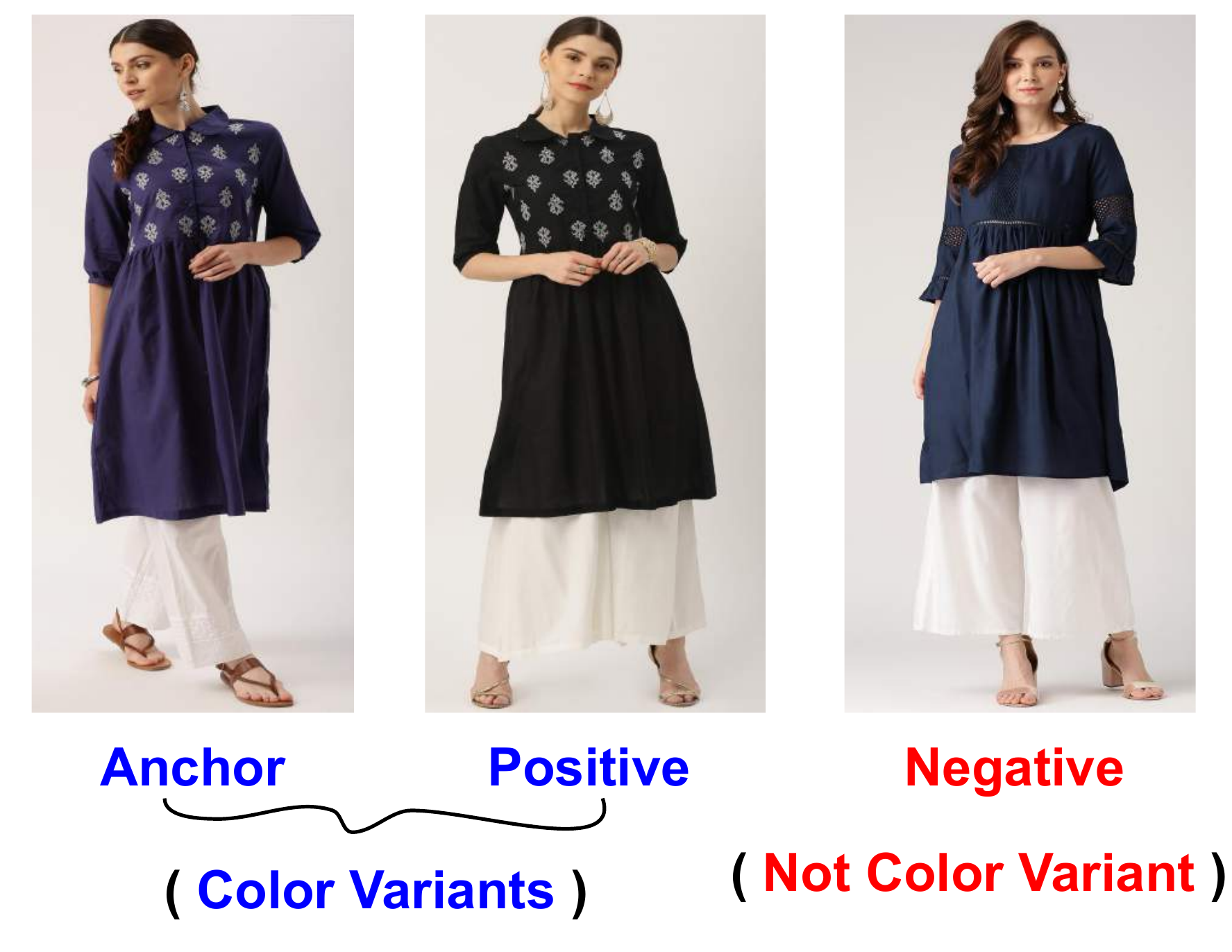}
  \caption{Illustration of the \textit{color variants identification} problem. The images belong to \url{www.myntra.com}.}
  \label{CV_illustration}
\end{figure}

In this paper, we address a very crucial problem in fashion e-commerce, namely, automated \textit{color variants identification}, i.e., identifying fashion products that match exactly in their design (or style), but only to differ in their color (Figure \ref{CV_illustration}). Our motivation to pick the use-case of color variants identification for fashion products comes from the following reasons: i) Fashion products top across all categories in online retail sales \cite{jagadeesh2014large}, ii) Most often users hesitate to buy a fashion product solely due to its color despite liking all other aspects of it. Providing more color options increases add-to-cart ratio, thereby generating more revenue, along with improved customer experience.
\begin{figure}[t]
  \centering
  \includegraphics[width=\columnwidth]{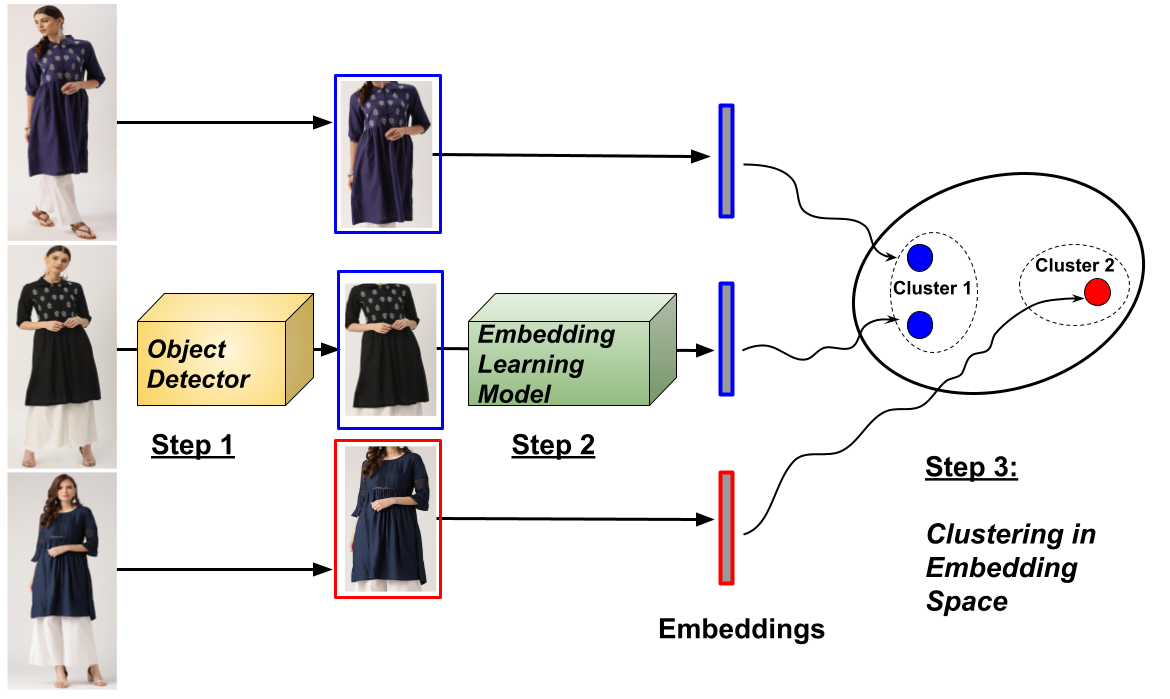}
  \caption{Proposed framework to address color variants identification in fashion e-commerce.}
  \label{CV_pipeline}
\end{figure}

At Myntra (\url{www.myntra.com}), a leading fashion e-commerce platform, we address this problem by leveraging deep visual Representation/ Embedding Learning. Our proposed framework, as shown in Figure \ref{CV_pipeline}, is generic in nature, consisting of an Embedding Learning model (Step 2), which is trained to obtain embeddings (i.e., representation of visual data in a vector space), in such a way that embeddings of color variant images are grouped together. Having obtained the embeddings, we can perform clustering (Step 3) on them to obtain the color variants (embeddings falling into the same cluster would correspond to images that are color variants). Also, as a typical Product Display Page (PDP) image in a fashion e-commerce platform usually contains multiple fashion products, we apply an object detector to localise our \textit{primary} fashion product of interest, as a preprocessing step (Step 1).

It should be noted that the \textit{main component in the pipeline is the Representation/ Embedding Learning model} in Step 2, and this is what we mainly focus in our paper. We discuss different strategies of training the Embedding Learning component. One way is to obtain manual annotations (class labels) indicating whether two product images are color variants to each other. These class labels are used to obtain triplet based constraints (details to be introduced later), consisting of an anchor, a positive and a negative (Figure \ref{CV_illustration}).

Despite performing well in practice, a key challenge in the real-world setting of fashion e-commerce is the large scale of data to be annotated, which often becomes infeasible, and tedious. \textit{Could we somehow get away with this annotation step ?}, this is what we asked ourselves. Interestingly, we noted that color variants of fashion products are in essence, manifestations of the same fashion \textit{stlye/ design} after applying color jittering, which is a widely popular image augmentation technique applied in visual Self-Supervised Learning (SSL). SSL finds image embeddings without requiring manual annotations! Thus, we go one step further, and try to answer the question of whether SSL could be employed in our use-case, for achieving comparable performance with that of a supervised model using manually labeled examples. For this, we evaluated a number of State-Of-The-Art (SOTA) SSL techniques from the recent literature, but found certain limitations in them, for our use-case. To address this, we make certain crafty modifications, and propose a novel SSL method as well.

Following are the \textbf{major contributions of the paper}:
\begin{enumerate}[noitemsep,nolistsep]
    \item A generic visual Representation Learning based framework to identify color variants among fashion products (to the best of our knowledge, studied as a research paper for the first time).
    \item A systematic study of a supervised method (with manual annotations), as well as existing SOTA SSL methods to train the embedding learning model.
    \item A novel contrastive loss based SSL method that focuses on parts of an object to identify color variants.
\end{enumerate}

\textbf{Related Work:}
The problem of visual embedding/ metric learning refers to that of obtaining vector representations/embeddings of images in a way that the embeddings of similar images are grouped together while moving away dissimilar ones. Several supervised \cite{circle_CVPR20,class_collapse_2020,gu2020symmetrical} and unsupervised \cite{dutta2020unsupervised,cao2019unsupervised,li2020unsupervised,SUML_AAAI20} approaches have been proposed in the recent literature. Contrastive learning \cite{simclr_20,moco_cvpr20,byol_20,simsiam_21}, a paradigm of visual SSL \cite{jing2020self}, groups together embeddings obtained from augmentations of the same image, without making use of manual annotations, as needed in Supervised approaches.

\section{Proposed Framework}
Our proposed framework to identify color variants, as already introduced in Figure \ref{CV_pipeline}, consists of the stages: i) Object Detection, ii) Embedding Learning and iii) Clustering. As the original input image usually consists of a human model wearing secondary fashion products as well, we perform object detection (using any off-the shelf object detector) to localise the primary fashion product of interest. The core component of our framework, i.e., the Embedding Learning model, can either be trained using a supervised method (when manual annotations are available), or a SSL method. The latter is useful when obtaining manual annotations is infeasible, at large scale.

We first discuss the supervised method, where we require a set of triplet images obtained from manually annotated examples (consisting of an anchor, positive, and negative, as shown in Figure \ref{CV_illustration}). The positive (p) is usually an image that consists of a product that is a color variant of the product contained in the anchor (a) image. The negative (n) is an image that consists of a product that is not a color variant of the products contained in the anchor and positive images. Let, the embeddings for the triplet of images be denoted as $(\Vec{x}_a, \Vec{x}_p, \Vec{x}_n)$. These triplets are used to train a supervised triplet loss based deep neural network model \cite{schroff2015facenet,veit2017conditional}, the objective of which is to bring the embeddings $\Vec{x}_a$ and $\Vec{x}_p$ closer, while moving away $\Vec{x}_n$. This is achieved by minimizing the following:
\begin{equation}
    \label{equation:tripletmarginrankingloss}
    \mathcal{L}_{triplet} = max(0, \lambda + \delta^2(\Vec{x}_a,\Vec{x}_p) - \delta^2(\Vec{x}_a,\Vec{x}_n) ),
    \end{equation}
such that $\delta^2(\Vec{x}_i,\Vec{x}_j)=\left \| \Vec{x}_i -\Vec{x}_j \right \|_2^2 $ denotes the squared Euclidean distance between the pair of examples $\Vec{x}_i$ and $\Vec{x}_j$, with $\left \| \Vec{x}_i \right \|_2^2$ being the squared $l_2$ norm of $\Vec{x}_i$, and $\lambda>0$ being a margin.

An appropriate clustering algorithm could be applied on the embeddings, such that an obtained cluster contains embeddings of images of products that are color variants to each other. However, the supervised model needs manual annotations which may be infeasible to obtain in large real-world datasets (such as those present in our platform). Thus, we explore SSL to identify color variants.

\subsection{SSL based Embedding Learning model}
\begin{figure}[t]
  \centering
  \includegraphics[width=\columnwidth]{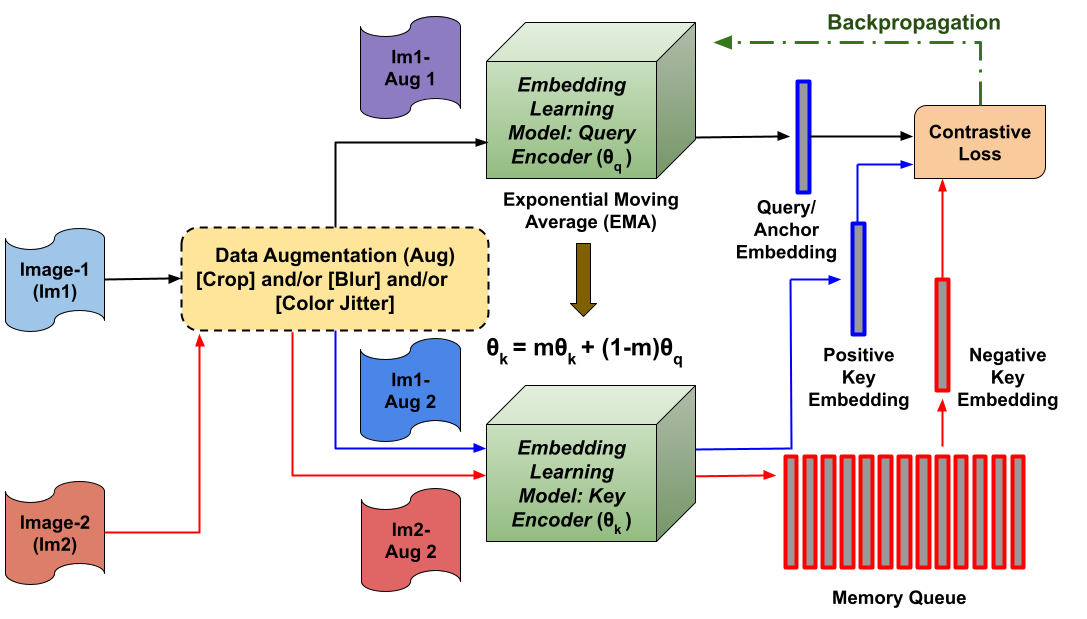}
  \caption{Contrastive SSL based Embedding Model.}
  \label{SSL_illus}
\end{figure}

\begin{figure}[t]
  \centering
  \includegraphics[width=0.9\columnwidth]{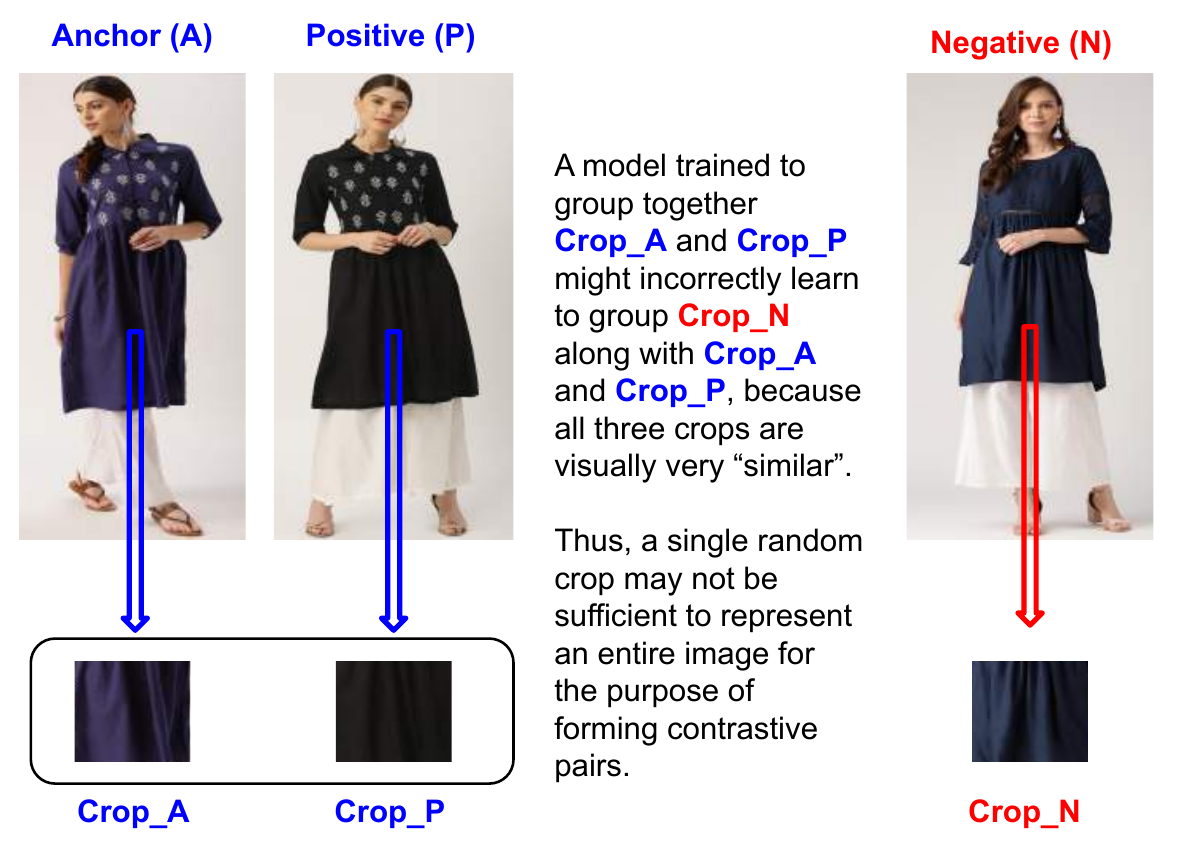}
  \caption{Drawbacks of using random crop to form contrastive pairs in our use-case.}
  \label{slice_motivation}
\end{figure}

SSL seeks to learn visual embeddings without requiring manual annotations. Figure \ref{SSL_illus} illustrates the contrastive loss based paradigm of SSL, using the SOTA approach called MOCOv2 \cite{moco_cvpr20}. Given an image (Im 1), two augmentations (Im 1-Aug 1 and Im 1-Aug2) are obtained, which are passed through the Embedding Learning model that is maintained as two branches: The Query Encoder and the Key encoder, with parameters $\theta_q$ and $\theta_k$ respectively. It should be noted that the Key encoder is simply a copy of the Query Encoder, and is maintained as a \textit{momentum encoder} obtained using an Exponential Moving Average (EMA) of the latter. The respective embeddings obtained for Im 1-Aug 1 and Im 1-Aug2 are treated as an anchor-positive pair. But, as we do not have manual annotations, to obtain the negative embedding (for avoiding a model collapse if just anchor and positive are pulled), one simply passes augmentation (Im 2-Aug 2) of another distinct image Im 2 (from within a mini-batch) through the key encoder. Additionally, to facilitate comparison with a large pool of negatives, another practice is to store the mini-batch embeddings obtained from the key encoder, in a separate memory module stored as a queue. The final embeddings of inference images are obtained by passing through the query encoder, which is the only branch that is updated via backpropagation (and not the key branch).

Other recent SOTA methods are built in a similar fashion as MOCOv2, with minor modifications. For instance, the BYOL \cite{byol_20} method does not make use of negatives (and hence no memory queue), but uses the EMA based update/ momentum encoder. The SimSiam method \cite{simsiam_21} neither uses negatives, nor momentum encoder. All these SSL methods (MOCOv2, BYOL and SimSiam) employ random cropping in the data augmentation pipeline. However, when we already have object detection in our pipeline, the standard random crop step used in existing SSL methods may actually miss out important regions of a fashion product, which might be crucial in identifying color variants (Figure \ref{slice_motivation}).
\begin{figure}[t]
  \centering
  \includegraphics[width=\columnwidth]{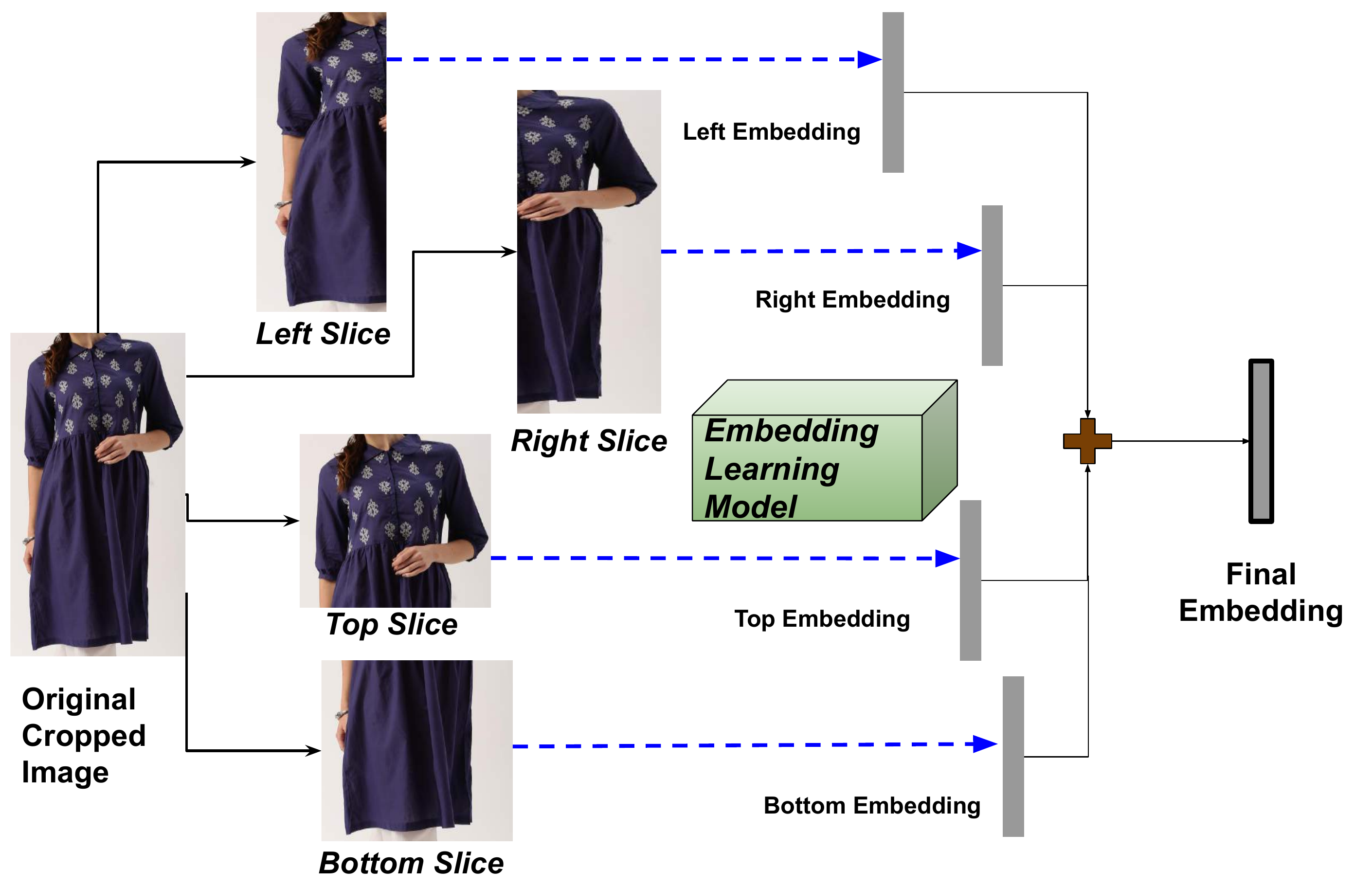}
  \caption{Illustration of our slicing based approach.}
  \label{PBCNet_illus}
\end{figure}

For this reason, we rather choose to propose a novel SSL variation that considers multiple slices/ patches (left, right, top and bottom) of the primary fashion object (after object detection), and simultaneously obtain embeddings for each of them. The final sum-pooled embedding\footnote{Rather than concatenating or averaging, we simply add them together to maintain simplicity of the model.} is then used to optimize a SSL based contrastive loss (Figure \ref{PBCNet_illus}). We make use of negative pairs in our method because we found the performance of methods that do not make use of negative pairs (eg, SimSiam \cite{simsiam_21}, BYOL \cite{byol_20}) to be sub-optimal in our use-case. We also found merits of momentum encoder and memory queue in our use-case, and thus include them in our method. Following is the Normalized Temperature-scaled cross entropy (NT-Xent) loss \cite{simclr_20} based objective of our method:
\begin{equation}
    \label{NTXent}
    \mathcal{L}_{\Vec{q}} = - \textrm{log} \frac{ \textrm{exp}( {\Vec{q}\Vec{k}_q} / \tau ) }{ \sum_{i=0}^K \textrm{exp}( {\Vec{q}\Vec{k}_i} / \tau ) }
\end{equation}
Here, $\Vec{q}=\sum_v \Vec{q}^{(v)}$, $\Vec{k}_i=\sum_v \Vec{k}_i^{(v)}, \forall i$. In (\ref{NTXent}), $\Vec{q}$ and $\Vec{k}_i$ respectively denote the \textit{final} embeddings obtained for a query and a key, which are essentially obtained by adding the embeddings obtained from across all the views, as denoted by the superscript $v$ for $\Vec{q}^{(v)}$ and $\Vec{k}_i^{(v)}$. Also, $\Vec{k}_q$ represents the positive key corresponding to a query $\Vec{q}$, $\tau$ denotes the temperature parameter, whereas $\textrm{exp}()$ and $\textrm{log}()$ respectively denote the exponential and logarithmic functions. $K$ in (\ref{NTXent}) denotes the size of the memory queue.

We call our method as \textbf{Patch-Based Contrastive Net (PBCNet)}. We conjecture that considering multiple patches i) do not leave things to chance (as in random crops), and provides a deterministic approach to obtain embeddings, ii) enables us to borrow more information (from other patches) to make a better decision on grouping a pair of similar embeddings. Our conjecture is supported not only by evidence of improved discriminative performances by the consideration of multiple fine patches of images, in literature \cite{wang2018learning}, but also by our experimental results.

\section{Experiments}
We evaluated the discussed methods on a large, challenging internal collection of images on our platform (\url{www.myntra.com}). Here, we report our results on a collection of Kurtas images. We used the exact same set to train the supervised (with labeled training data, obtained by our in-house team) and self-supervised methods (without labeled training data) for a fair comparison. For inferencing, we use 6 dataset splits (based on brand, gender, MRP), referred to as Data 1-6.
\textbf{Performance metrics used:} To evaluate the methods, we made use of the following performance metrics: i) Color Group Accuracy (\textbf{CGacc}), an internal metric with business relevance, that reflects the \textit{precision}, ii) Adjusted Random Index (\textbf{ARI}), iii) Fowlkes-Mallows Score (\textbf{FMS}), and iv) Clustering Score (\textbf{CScore}), computed as: $CScore = \frac{2.ARI.FMS}{ARI+FMS}$. While we use CGacc to compare the methods for all the datasets (\textit{Data 1-6}), the remaining metrics are reported only for \textit{Data 4-6}, where we obtain the ground-truth labels for evaluation. All the performance metrics take values in the range $[0,1]$, where a higher value indicates a better performance.

\textbf{Methods Compared}: The focus of our experiments is to evaluate the overall performance by using different embedding learning methods (while fixing the object detector and the clustering algorithm). We compare the supervised triplet loss based model, and our proposed PBCNet method, against the following SOTA SSL methods: \textbf{SimSiam} \cite{simsiam_21}, \textbf{BYOL} \cite{byol_20}, and \textbf{MOCOv2} \cite{moco_cvpr20}.

We implemented all the methods in PyTorch. For all the compared methods, we fix a ResNet34 \cite{ResNet} as a base encoder with  $224 \times 224$ image resizing, and train all the models for a fixed number of 30 epochs, for a fair comparison. The number of epochs was fixed based on observations on the supervised model, to avoid overfitting. For the purpose of object detection, we made use of YOLOv2 \cite{yolov2}, and for the task of clustering the embeddings, we made use of the Agglomerative Clustering algorithm with Ward's method for merging. In all cases, the 512-dimensional embeddings used for clustering are obtained using the avgpool layer of the trained encoder. A margin of 0.2 has been used in the triplet loss for training the supervised model.

\subsection{Systematic Study of SSL for our use-case}

\begin{table}[!t]
\centering
\resizebox{0.9\columnwidth}{!}{%
\begin{tabular}{|c|c|cc|cc|}
\hline
Dataset                 & Metric & \begin{tabular}[c]{@{}c@{}}SimSiam\_v0\\ (w/o \\ norm)\end{tabular} & SimSiam\_v0   & \begin{tabular}[c]{@{}c@{}}SimSiam\_v1\\ (w/o \\ norm)\end{tabular} & SimSiam\_v1   \\ \hline
Data 1                  & CGacc  & 0.5                                                                          & 0.5           & \textbf{1}                                                                   & 0.67          \\ \hline
Data 2                  & CGacc  & 0.5                                                                          & \textbf{0.67} & 0                                                                            & \textbf{0.25} \\ \hline
Data 3                  & CGacc  & 0                                                                            & \textbf{0.4}  & 0                                                                            & \textbf{0.33} \\ \hline
\end{tabular}%
}
\caption{Effect of Embedding Normalization.}
\label{simsiam_normalization}
\end{table}   
We now perform a systematic study of the typical aspects associated with SSL, especially for our particular task of color variants identification. For this purpose, we make use of a single Table \ref{results_all}, where we provide the comparison of various SSL methods, including ours.

\textbf{Effect of Data augmentation on our task:} Firstly, for illustrating the effect of different data augmentations, we consider two variants of SimSiam (for its simplicity and strength): i) SimSiam\_v0: A version of SimSiam, where we used the entire original image as the query, and a color jittered image as the positive, and with a batch size of 12, and ii) SimSiam\_v1: SimSiam with standard SSL \cite{simclr_20} augmentations (\texttt{ColorJitter}, \texttt{RandomGrayscale}, \texttt{RandomHorizontalFlip}, \texttt{GaussianBlur} and \texttt{RandomResizedCrop}), and a batch size of 12. For all cases, the following architecture has been used for SimSiam: \texttt{Encoder}\{ ResNet34 $\rightarrow$ (avgpool) $\rightarrow$ \texttt{ProjectorMLP}(512$\rightarrow$4096$\rightarrow$123) \}$\rightarrow$\texttt{PredictorMLP}(123$\rightarrow$4096$\rightarrow$123).

From Table \ref{results_all}, we observed that considering standard augmentations with random crops leads to a better performance than that of SimSiam\_v0. Thus, for all other self-supervised baselines, i.e., BYOL and MOCOv2 we make use of standard augmentations with random crops. However, we later show that our proposed way of considering multiple patches in PBCNet leads to a better performance.

\textbf{Effect of l2 normalization for our task:} Table \ref{simsiam_normalization} shows the effect of performing $l2$ normalization on the embeddings obtained using SimSiam. We found that without using any normalization, in some cases (eg, Data 2-3) there are no true color variant groups out of the detected clusters (i.e., zero precision), and hence the performance metrics become zero. Thus, for all our later experiments, we make use of $l2$ normalization on the embeddings as a de facto standard, for all the methods.
\begin{table*}[t]
\centering
\resizebox{0.7\linewidth}{!}{%
\begin{tabular}{|c|c|c|ccccc|c|}
\hline
\multicolumn{2}{|c|}{}            & {\color[HTML]{3531FF} \textbf{Supervised}}  & \multicolumn{6}{c|}{Self-Supervised}                                                    \\ \hline
Dataset                  & Metric & {\color[HTML]{3531FF} \textbf{Triplet Net}} & SimSiam\_v0 & SimSiam\_v1 & SimSiam\_v2 & BYOL  & MOCOv2       & \textbf{PBCNet (Ours)} \\ \hline
Data 1                   & CGacc  & {\color[HTML]{3531FF} \textbf{0.67}}        & 0.5         & 0.67        & 1           & 0.5   & 1            & \textbf{1}             \\ \hline
Data 2                   & CGacc  & {\color[HTML]{3531FF} \textbf{1}}           & 0.67        & 0.25        & 1           & 0.75  & \textbf{0.8} & 0.75                   \\ \hline
Data 3                   & CGacc  & {\color[HTML]{3531FF} \textbf{0.75}}        & 0.4         & 0.33        & 0           & 0.5   & 0.5          & \textbf{0.6}           \\ \hline
                         & CGacc  & {\color[HTML]{3531FF} \textbf{0.67}}        & 0.4         & 0.5         & 0.5         & 0.5   & \textbf{1}   & 0.85                   \\
                         & ARI    & {\color[HTML]{3531FF} \textbf{0.69}}        & 0.09        & 0.15        & 0.12        & 0.27  & 0.66         & \textbf{0.75}          \\
                         & FMS    & {\color[HTML]{3531FF} \textbf{0.71}}        & 0.15        & 0.22        & 0.20        & 0.30  & 0.71         & \textbf{0.76}          \\
\multirow{-4}{*}{Data 4} & CScore & {\color[HTML]{3531FF} \textbf{0.700}}       & 0.110       & 0.182       & 0.152       & 0.281 & 0.680        & \textbf{0.756}         \\ \hline
                         & CGacc  & {\color[HTML]{3531FF} \textbf{1}}           & 0           & 0.5         & 0.33        & 0.5   & 1            & \textbf{1}             \\
                         & ARI    & {\color[HTML]{3531FF} \textbf{1}}           & 0           & 0.09        & 0.28        & 0.64  & 1            & \textbf{1}             \\
                         & FMS    & {\color[HTML]{3531FF} \textbf{1}}           & 0.22        & 0.30        & 0.45        & 0.71  & 1            & \textbf{1}             \\
\multirow{-4}{*}{Data 5} & CScore & {\color[HTML]{3531FF} \textbf{1}}           & 0           & 0.135       & 0.341       & 0.674 & 1            & \textbf{1}             \\ \hline
                         & CGacc  & {\color[HTML]{3531FF} \textbf{0.83}}        & 0.5         & 0.5         & 0.8         & 0.6   & 1            & \textbf{1}             \\
                         & ARI    & {\color[HTML]{3531FF} \textbf{0.44}}        & 0.07        & 0.06        & 0.04        & 0.20  & 0.58         & \textbf{0.79}          \\
                         & FMS    & {\color[HTML]{3531FF} \textbf{0.49}}        & 0.12        & 0.17        & 0.13        & 0.24  & 0.64         & \textbf{0.80}          \\
\multirow{-4}{*}{Data 6} & CScore & {\color[HTML]{3531FF} \textbf{0.466}}       & 0.089       & 0.090       & 0.063       & 0.214 & 0.610        & \textbf{0.796}         \\ \hline
\end{tabular}%
}
\caption{Comparison of our proposed method against the supervised and state-of-the-art SSL baselines, across all the datasets.}
\label{results_all}
\end{table*}

\textbf{Effect of Batch Size and Momentum Encoding in SSL for our task:}
For studying the effect of batch size in SSL for our task, we introduce a third variant of SimSiam, i.e., SimSiam\_v2: This is essentially SimSiam\_v1 with a batch size of 128. We then consider SimSiam\_v1, SimSiam\_v2 and BYOL, where the first makes use of a batch size of 12, while the others make use of batch sizes of 128. We observed that a larger batch size usually leads to a better performance. This is observed from Table \ref{results_all}, by the higher values of performance metrics in the columns for SimSiam\_v2 and BYOL (vs SimSiam\_v1). Additionally, we noted that the momentum encoder used in BYOL causes a further boost in the performance, as observed in its superior performance as compared to SimSiam\_v2 that has the same batch size. It should be noted that except for the momentum encoder, the rest of the architecture and augmentations used in BYOL are exactly the same as in SimSiam. We observed that increasing the batch size in SimSiam does not drastically or consistently improve its performance, something which its authors also noticed \cite{simsiam_21}.

\textbf{Effect of Memory Queue in SSL for our task:} 
We also inspect the effect of an extra memory module/queue being used to facilitate the comparisons with a large number of negative examples. In particular, we make use of the MOCOv2 method with the following settings: i) queue size of 5k, ii) temperature parameter of 0.05, iii) a MLP (512$\rightarrow$4096$\rightarrow$relu$\rightarrow$123) added after the avgpool layer of the ResNet34, iv) SGD for updating the query encoder, with learning rate of 0.001, momentum of 0.9, weight decay of $1e^{-6}$, and v) value of 0.999 for $m$ in the momentum update. It is observed from Table \ref{results_all}, by the columns of MOCOv2 and BYOL, that the performance of the former is superior. As BYOL does not use a memory module, but MOCOv2 does, we conclude that using a separate memory module significantly boosts the performance of SSL in our task. Motivated by our observations so far, we choose to employ both momentum encoding and memory module in our proposed PBCNet method.

\subsection{Comparison of PBCNet against the state-of-the-art}
In Table \ref{results_all}, we provide the comparison of our proposed SSL method \textbf{PBCNet} against the SSL SOTA baselines and the supervised baseline across all the datasets. It should be noted that in Table \ref{results_all}, any performance gains for a specific method is due to the intrinsic nature of the same, and not because of a particular hyperparameter setting. This is because we report the \textit{best performance for each method} after adequate tuning of the clustering distance threshold and other parameters, and not just their default hyperparameters.

Following are the configurations that we have used in our \textbf{PBCNet} method: i) memory module size of 5k, ii) temperature parameter of 0.05, iii) the FC layer after the avgpool layer of the ResNet34 was removed, iv) SGD for updating the query encoder, with learning rate of 0.001, momentum of 0.9, weight decay of $1e^{-6}$, and v) value of 0.999 for $m$ in the momentum update.

We made use of a batch size of $32 (=128/4)$ as we have to store tensors for each of the 4 slices simultaneously for each mini-batch (we used a batch size of 128 for the other methods). For data augmentation, we first apply a color distortion in the following order: i) \texttt{ColorJitter(0.8 * s, 0.8 * s, 0.8 * s, 0.2 * s)} with s=1, p=0.8, ii) \texttt{RandomGrayscale} with p=0.2, iii) \texttt{GaussianBlur((3, 3), (1.0, 2.0))} with p=0.5. After the color distortion, we apply our slicing technique. For the second image (positive/negative) we apply the same series of transformations.
\begin{figure}[t]
  \centering
  \includegraphics[width=\columnwidth]{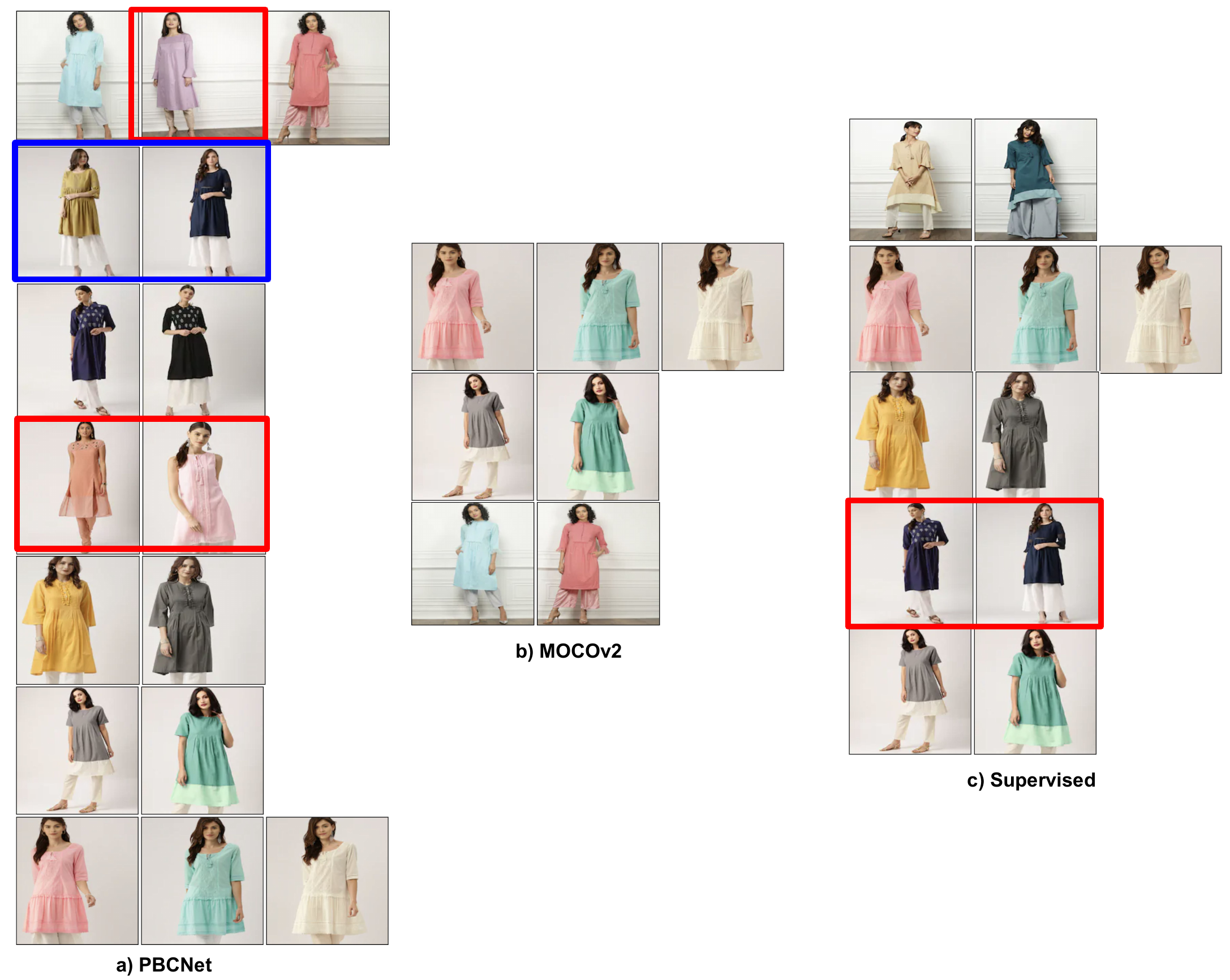}
  \caption{Qualitative comparison of color variants groups obtained using our PBCNet method (left column), MOCOv2 (middle column) and the supervised baseline (right column), on \textit{Data 4}.}
  \label{Data4_quali}
\end{figure}

From Table \ref{results_all}, it is clear that the SimSiam method despite its strong claims of not using any negative pairs, nor momentum encoder nor large batches, performs poorly as compared to our supervised method (shown in bold blue color). The BYOL method which also does not make use of negative pairs, performs better than the SimSiam method in our use case, by virtue of its momentum encoder.

Among all compared SSL baselines, it is the MOCOv2 method that performs the best. This is due to the reason of the memory queue that facilitates the comparison with a large number of negative examples. This shows that the importance of considering negative pairs still holds true, especially for challenging use-cases like the one considered in the paper. However, our proposed self-supervised method \textbf{PBCNet} clearly outperforms all the baselines. The fact that it outperforms MOCOv2 can be attributed to the patch-based slicing used, which is the only different component in our method in comparison to MOCOv2 that uses random crop. Another interesting thing that we observed is the fact that despite using much lesser batch size of 32, our method outperforms the baselines. In a way, we were able to extract and leverage more information by virtue of the slicing (by borrowing information from the other patches simultaneously), even with smaller batches.

We also noticed that the supervised baseline performs quite good in our task, even without any data augmentation pipeline as used in the SSL methods. However, by virtue of the large memory queue, and the data augmentations like color jitter, which are pretty relevant to the task of color variants identification, stronger SSL methods like MOCOv2 and PBCNet are in fact capable of achieving comparable performance against the supervised baseline. Having said that, if we do not have adequate labeled data in the first place, we cannot even use supervised learning. Hence, enabling data augmentations and slicing strategy in the supervised model has not been focused, because the necessity of our approach comes from the issue of addressing the lack of labeled data, and not to improve the performance of supervised learning (which any how is label dependent).

\textbf{Effect of Clustering:}
In Table \ref{effect_cluster}, we report the performances obtained by varying the clustering algorithm to group embeddings obtained by different SSL methods, on \textit{Data 4-6}. We picked the Agglomerative, DBSCAN and Affinity Propagation clustering techniques that do not require the number of clusters as input parameter (which is difficult to obtain in our use-case). In general, we observed that the Agglomerative clustering technique leads to a better performance in our use-case. Also, for a fixed clustering approach, using embeddings obtained by our PBCNet method usually leads to a better performance.
\begin{table}[t]
\centering
\resizebox{0.8\columnwidth}{!}{%
\begin{tabular}{|c|c|cc|cc|cc|}
\hline
\multicolumn{2}{|c|}{Dataset} & \multicolumn{2}{c|}{\textbf{Data 4}} & \multicolumn{2}{c}{\textbf{Data 5}} & \multicolumn{2}{c|}{\textbf{Data 6}} \\ \hline
Method           & Clustering & \textbf{ARI}      & \textbf{FMS}     & \textbf{ARI}     & \textbf{FMS}     & \textbf{ARI}      & \textbf{FMS}     \\ \hline
\textbf{PBCNet}  & Agglo      & \textbf{0.75}     & \textbf{0.76}    & \textbf{1.00}    & \textbf{1.00}    & \textbf{0.79}     & \textbf{0.80}    \\
                 & DBSCAN     & 0.66              & 0.71             & 1.00             & 1.00             & 0.66              & 0.71             \\
                 & Affinity   & 0.30              & 0.42             & 0.22             & 0.41             & 0.24              & 0.37             \\ \hline
\textbf{MOCOv2}  & Agglo      & \textbf{0.66}     & \textbf{0.71}    & \textbf{1.00}    & \textbf{1.00}    & \textbf{0.58}     & \textbf{0.64}    \\
                 & DBSCAN     & 0.66              & 0.71             & 1.00             & 1.00             & 0.37              & 0.40             \\
                 & Affinity   & 0.20              & 0.32             & 0.04             & 0.26             & 0.25              & 0.41             \\ \hline
\textbf{BYOL}    & Agglo      & \textbf{0.27}     & \textbf{0.30}    & \textbf{0.64}    & \textbf{0.71}    & \textbf{0.20}     & \textbf{0.24}    \\
                 & DBSCAN     & 0.17              & 0.28             & 0.64             & 0.71             & 0.02              & 0.24             \\
                 & Affinity   & 0.03              & 0.14             & 0.28             & 0.45             & 0.01              & 0.11             \\ \hline
\end{tabular}%
}
\caption{Effect of the clustering technique used.}
\label{effect_cluster}
\end{table}

\begin{figure}[t]
  \centering
  \includegraphics[width=0.85\columnwidth]{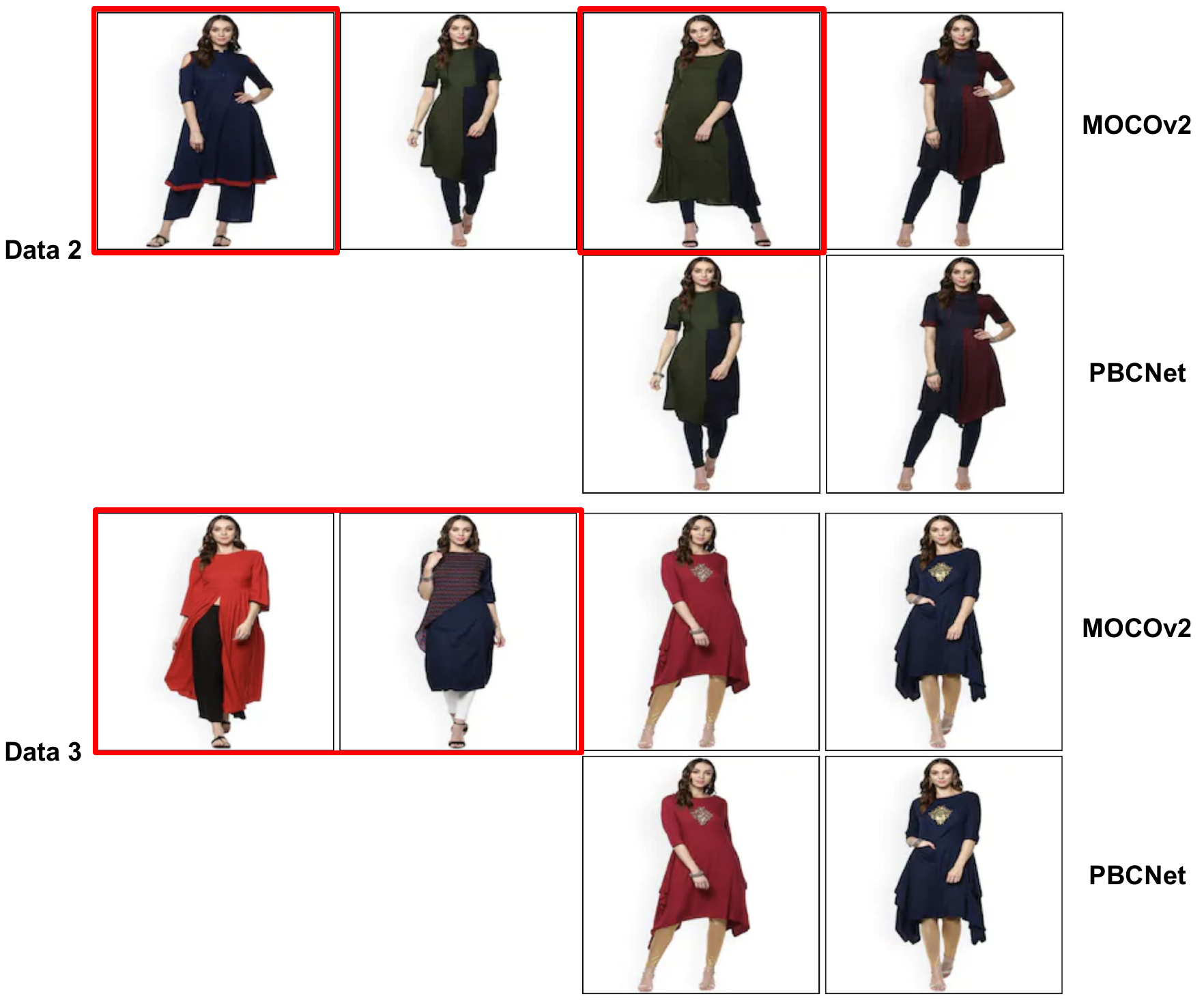}
  \caption{A few groups obtained on \textit{Data 2 \& 3} using MOCOv2 have false positives (shown in red box), while our PBCNet method does not yield such groups.}
  \label{PBCNet_vs_MOCOv2_quali}
\end{figure}

\begin{figure}[t]
  \centering
  \includegraphics[width=\columnwidth]{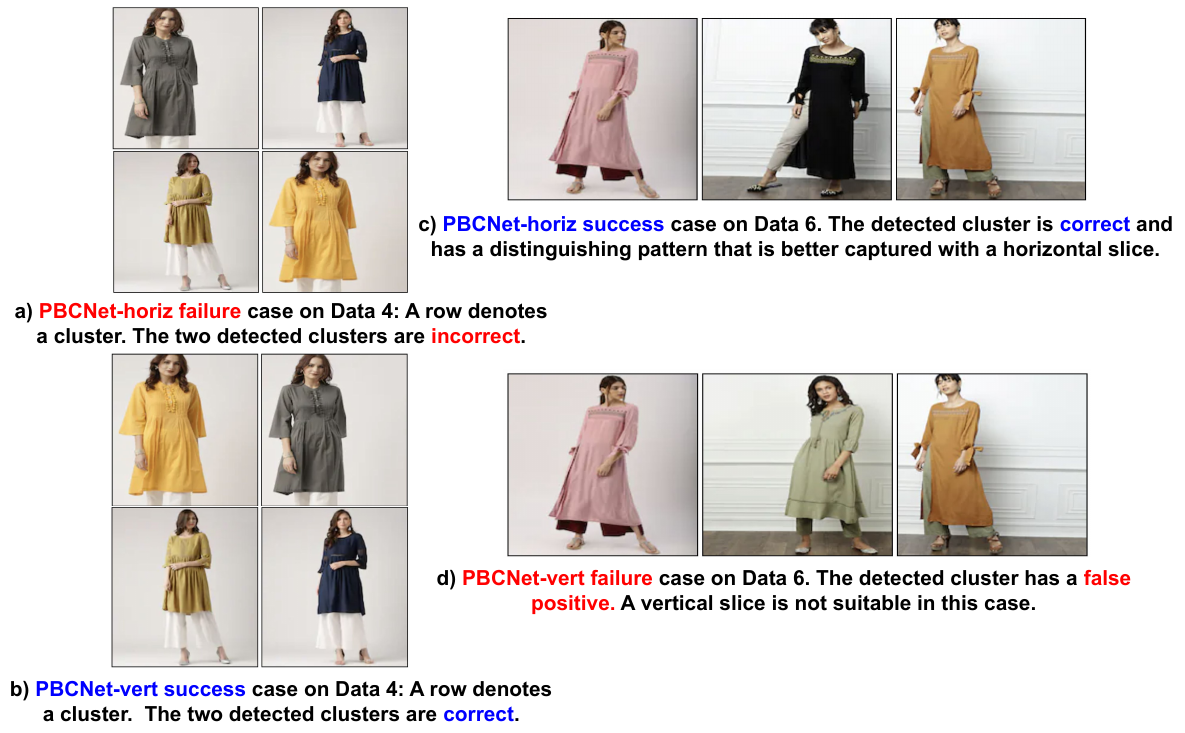}
  \caption{Trade off between vertical and horizontal slicing.}
  \label{slicing_tradeoff}
\end{figure}

\begin{table}[t]
\centering
\resizebox{\columnwidth}{!}{%
\begin{tabular}{|c|c|c|c|ccc|ccc|ccc|}
\hline
Dataset         & Data 1 & Data 2     & Data 3       & \multicolumn{3}{c|}{Data 4}                                                             & \multicolumn{3}{c|}{Data 5} & \multicolumn{3}{c|}{Data 6}                                   \\ \hline
Method          & CGacc  & CGacc      & CGacc        & CGacc         & ARI                                & FMS                                & CGacc     & ARI    & FMS    & CGacc & ARI                       & FMS                       \\ \hline
PBCNet-horiz    & 1      & \textbf{1} & 0.5          & 0.66          & 0.65                               & 0.67                               & 1         & 1      & 1      & 1     & \textbf{0.81}             & \textbf{0.82}             \\
PBCNet-vert     & 1      & 0.6        & 0.6          & \textbf{1} & \textbf{0.88} & \textbf{0.89} & 1         & 1      & 1      & 0.83  & 0.48 & 0.50 \\ \hline
\textbf{PBCNet} & 1      & 0.75       & \textbf{0.6} & 0.85          & 0.75                               & 0.76          & 1         & 1      & 1      & 1     & 0.79 & 0.80 \\ \hline
\end{tabular}%
}
\caption{Effect of Slicing on PBCNet}
\label{effect_slicing}
\end{table}
\textbf{Qualitative results:} Sample qualitative comparisons of color variants groups obtained on \textit{Data 4} using our PBCNet method, MOCOv2 and the supervised baseline are provided in Figure \ref{Data4_quali}. Each of the rows for a column corresponding to a method represents a detected color variants cluster for the considered method. A row has been marked with a red box if the entire cluster contains images that are not color variants to each other. A single image is marked with a red box if it is the only incorrect image, while rest of the images are color variants. We observed that our method not only detects clusters with higher precision (which MOCOv2 does as well), but it also has a higher recall, which is comparable to the supervised method. We also make use of a blue box to show a detected color group by our method which contains images that are color variants, but are difficult to be identified at a first glance, even for humans.

Additionally, Figure \ref{PBCNet_vs_MOCOv2_quali} shows a few color groups identified in the datasets \textit{Data 2 \& 3} using MOCOv2 and our PBCNet. We observed that MOCOv2 detected groups with false positives, while our PBCNet method did not. This could happen because when a random crop is obtained by MOCOv2, it need not necessarily be from a \textit{distinctive} region of an apparel that helps to identify its color variant  (eg, in Figure \ref{PBCNet_vs_MOCOv2_quali}, the bent line like pattern separating the colored and black region of the apparels of Data 2, and the diamond like shape in the apparels of Data 3). We argue that a random crop might have arrived from such a \textit{distinctive} region given that the size of the crop is made larger, etc. But that still leaves things to random chance. On the other hand, our slicing technique being deterministic in nature, \textit{guarantees} that all the regions of an object \textit{would} be captured. We would also like to mention that our slicing approach is agnostic to the fashion apparel type, i.e, the same is easily applicable for any fashion article type (Tops, Shirts, Shoes, Trousers, Track pants, Sweatshirts, etc). In fact, this is how a human identifies color variants as well, by looking at the article along both horizontal and vertical directions, to identify distinctive patterns. Even humans cannot identify an object if we restrict our vision to only a particular small crop.

\textbf{Effect of the slicing}: We also study 2 variants of our PBCNet method: i) PBCNet-horiz (computing an embedding only by considering the top and bottom slices), and ii) PBCNet-vert (computing an embedding using only the left and right slices). The results are shown in Table \ref{effect_slicing}. In \textit{Data 4}, PBCNet-vert performs better than PBCNet-horiz, and in \textit{Data 6}, PBCNet-horiz performs better than PBCNet-vert (significantly). The performance of the two versions is also illustrated in Figure \ref{slicing_tradeoff}. We observed that a single slicing do not work in all scenarios, especially for apparels.

Although the horizontal slicing is quite competitive, it may be beneficial to consider the vertical slices as well. This is observed by the drop in performance of PBCNet-horiz in \textit{Data 3-4} (vs PBCNet). This is because some garments may contain distinguishing patterns that may be better interpreted only by viewing vertically, for example, printed texts (say, \textit{adidas} written vertically), floral patterns etc. In such cases, simply considering horizontal slices may actually split/ disrupt the vertical information. It may also happen that mixing of slicing introduces some form of redundancy, as observed by the occasional drop in the performance of PBCNet when compared to PBCNet-horiz (on \textit{Data 6}) and PBCNet-vert (on \textit{Data 4}). However, on average PBCNet leads to an overall consistent and competitive performance, while avoiding drastically fluctuating improvements or failures. We suggest considering both the directions of slicing, so that they could collectively represent all necessary and distinguishing patterns, and if one slicing misses some important information, the other could compensate for it.

\section{Conclusions}
In this paper, a generic visual Representation Learning based framework to identify color variants in fashion e-commerce has been studied (to the best of our knowledge, for the first time). A systematic study of a supervised method (with manual annotations), as well as existing SOTA SSL methods to train the embedding learning component of our framework has been conducted. A novel contrastive loss based SSL method that focuses on parts of an object to identify color variants, has also been proposed.

\section{Acknowledgement}
We would like to thank Dr Ravindra Babu Tallamraju for his support, feedback, and in being a source of inspiration and encouragement throughout the project.

\bibliography{PBCNet_IAAI22}

\end{document}